\documentclass[letterpaper]{article} 
\usepackage{aaai2026}  
\usepackage{times}  
\usepackage{helvet}  
\usepackage{courier}  
\usepackage[hyphens]{url}  
\usepackage{graphicx} 
\usepackage{natbib}  
\usepackage{caption} 
\usepackage{algorithm}
\usepackage{algorithmic}

\usepackage{newfloat}
\usepackage{listings}
\DeclareCaptionStyle{ruled}{labelfont=normalfont,labelsep=colon,strut=off} 
\floatstyle{ruled}
\newfloat{listing}{tb}{lst}{}
\floatname{listing}{Listing}
\usepackage{booktabs}
\usepackage{multirow}
\usepackage{adjustbox}
\usepackage[table,dvipsnames]{xcolor}
\definecolor{lightblue}{rgb}{0.84,0.91,0.97}
\usepackage{xspace}
\usepackage{outlines}
\makeatletter
\DeclareRobustCommand\onedot{\futurelet\@let@token\@onedot}
\def\@onedot{\ifx\@let@token.\else.\null\fi\xspace}
\def\eg{\emph{e.g}\onedot} 
\def\ie{\emph{i.e}\onedot} 
\usepackage{amsmath}
\usepackage{amsfonts}
\newcommand{\q}[1]{`#1'}
\newcommand{\dq}[1]{``#1''}

\usepackage{subcaption}

\title{TabFlash: Efficient Table Understanding with \\ Progressive Question Conditioning and Token Focusing}
\author{
    Jongha Kim\textsuperscript{\rm 1}, Minseong Bae\textsuperscript{\rm 2}\thanks{This work was conducted at Korea University.}, Sanghyeok Lee\textsuperscript{\rm 2}\footnotemark[1], Jinsung Yoon\textsuperscript{\rm 3}, Hyunwoo J. Kim\textsuperscript{\rm 2}\thanks{Corresponding Author.}
}
\affiliations{
    \textsuperscript{\rm 1}Korea University,
    \textsuperscript{\rm 2}KAIST,
    \textsuperscript{\rm 3}Google Cloud AI

    jonghakim@korea.ac.kr, \{bms2002,sanghyeoklee,hyunwoojkim\}@kaist.ac.kr, jinsungyoon@google.com
}

\usepackage{bibentry}

\begin{document}

\maketitle

\begin{abstract}
Table images present unique challenges for effective and efficient understanding due to the need for question-specific focus and the presence of redundant background regions.
Existing Multimodal Large Language Model (MLLM) approaches often overlook these characteristics, resulting in uninformative and redundant visual representations.
To address these issues, we aim to generate visual features that are both informative and compact to improve table understanding.
We first propose progressive question conditioning, which injects the question into Vision Transformer layers with gradually increasing frequency, considering each layer’s capacity to handle additional information, to generate question-aware visual features.
To reduce redundancy, we introduce a pruning strategy that discards background tokens, thereby improving efficiency.
To mitigate information loss from pruning, we further propose token focusing, a training strategy that encourages the model to concentrate essential information in the retained tokens.
By combining these approaches, we present TabFlash, an efficient and effective MLLM for table understanding.
TabFlash achieves state-of-the-art performance, outperforming both open-source and proprietary MLLMs, while requiring 27\% less FLOPs and 30\% less memory usage compared to the second-best MLLM. 
\end{abstract}

\begin{links}
    \link{Code}{https://github.com/mlvlab/TabFlash}
\end{links}

\section{Introduction}
Table data is a vital information source, widely used to organize and communicate structured knowledge across diverse domains.
With the recent success of Multimodal Large Language Models (MLLMs)~\cite{alayrac2022flamingo,li2023blip,liu2023visual,chen2024expanding}, MLLM-based methods have gained popularity for table image understanding~\cite{zheng2024multimodal,zhao2024tabpedia,zhou2025syntab}.
While these methods demonstrate the potential of MLLMs in table understanding, they often overlook the unique challenges posed by table images.

\begin{figure}[t!]
\begin{center}
\includegraphics[width=1.0\linewidth]{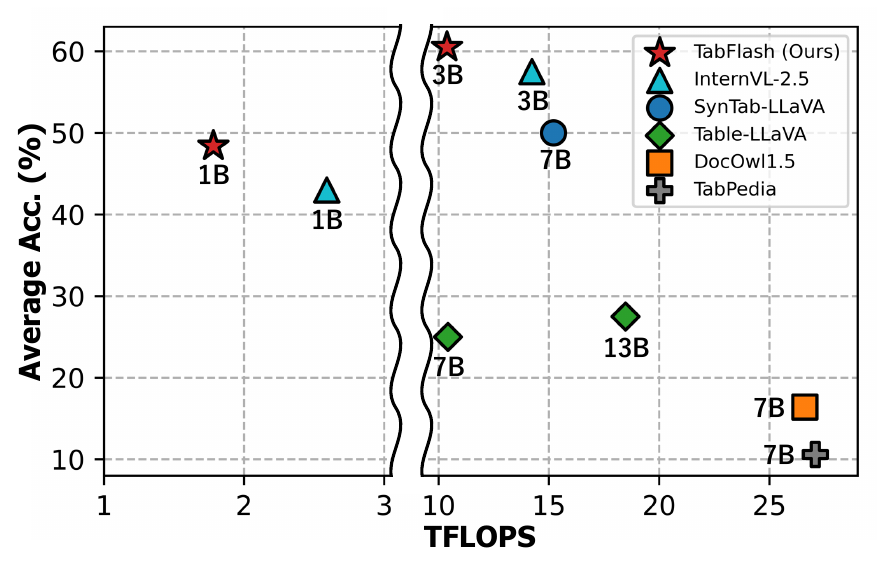}
\end{center}
\caption{
\textbf{Performance-cost comparison.}
TFLOPs (x-axis) and average accuracy on 7 benchmarks (y-axis) are plotted.
We propose TabFlash, an efficient MLLM with superior table understanding capability (Tab.~\ref{tab:main_table}) with significantly lower computational cost and GPU memory usage (Tab.~\ref{tab:cost_comparison}).
}
\label{fig:performance_cost_fig}
\end{figure}
Unlike natural images, table images require focused attention on localized regions relevant to a specific question, as the majority of the image content is typically irrelevant to the target task.
In addition, they often include substantial redundancy, such as empty or background areas.
Existing MLLMs struggle to handle these characteristics effectively, leading to the generation of uninformative and redundant visual representations.
This not only degrades performance but also incurs high inference costs.
To address these limitations, we aim to produce compact and informative visual representations tailored to the unique structure of table images.

As a solution, we first introduce progressive question conditioning, a strategy that injects question information into the Vision Transformer (ViT)~\cite{dosovitskiy2020image} to produce more informative visual features.
Specifically, we embed the question into ViT layers with gradually increasing conditioning frequencies.
In this design, early layers are conditioned less frequently, while later layers are conditioned more often.
This approach is based on the observation that early layers are more sensitive and unstable, whereas later layers are more stable and better suited to incorporate additional information~\cite{dosovitskiy2020image,raghu2021vision}.
By adjusting the conditioning frequency to match each layer’s capacity, progressive conditioning enables stable and effective integration of question context.
As a result, the ViT generates visual features that are more closely aligned with the question, improving their informativeness.

In addition, we propose a background pruning strategy to generate more compact visual features.
Previous MLLM-based approaches~\cite{chen2024expanding,zhao2024tabpedia,zhou2025syntab} often provide a large number of visual tokens to the language model (LM), in some cases exceeding 3,000 tokens.
Given the quadratic complexity of an LM concerning input tokens, this leads to a significant computational burden.
However, much of a table image consists of redundant background pixels.
We observe that the $L_2$ norm of output tokens from the ViT can serve as an effective signal to identify background regions, with tokens having lower norms typically corresponding to background areas (Fig.~\ref{fig:l2_norm_visualization_fig}).
Based on this observation, we introduce a token pruning strategy that removes low-norm tokens and passes only the retained tokens to the language model, reducing computation.

Still, we observe that a simple token pruning leads to significant performance loss.
Through observation, we identify that useful information for answering the question is still present in tokens to be discarded.
As a result, pruning those tokens without additional guidance causes information loss (Sec.~\ref{sec:token_focusing_analysis}).
To address this issue, we propose token focusing, a training strategy that encourages the model to concentrate important information within the tokens that will be retained.
Token focusing guides the model to produce correct answers based on the retained tokens, while discouraging correct predictions when only using the tokens designated for removal.
By doing so, the model is promoted to store important information only on tokens to be retained during inference, thereby minimizing information loss.

Combining the proposed methods, we introduce TabFlash, an efficient MLLM for table understanding.
These components work together to produce visual features that are both informative and compact, resulting in substantial improvements in effectiveness and efficiency.
As shown in Fig.~\ref{fig:performance_cost_fig}, TabFlash achieves state-of-the-art performance while requiring significantly lower computational resources.
Remarkably, TabFlash even surpasses proprietary models such as GPT-4o and Gemini 2.5 Pro, highlighting both the challenge of table image understanding and the strength of our approach.
In sum, our contributions are threefold:
\begin{itemize}
    \item We propose progressive question conditioning, injecting questions to ViT layers in a progressively increasing frequency, thus obtaining question-aware visual features.
    \item We prune visual tokens based on their $L_2$ norm for efficiency.
    We also introduce token focusing, a training strategy that enforces information concentration on tokens to be retained, minimizing the information loss by pruning.
    \item
    We introduce TabFlash, an efficient MLLM achieving state-of-the-art results.
    TabFlash outperforms the second-best open-source model by 3 points while requiring 27\% less FLOPs and 30\% less memory.
    It also surpasses proprietary models, such as GPT-4o and Gemini 2.5 Pro.
\end{itemize}

\section{Related Works}
\noindent\textbf{Table understanding with MLLMs.}
Table understanding involves interpreting and reasoning over tabular data, including tasks like question answering~\cite{pasupat2015compositional,zhu2021tat,lu2022dynamic}, fact verification~\cite{chen2019tabfact,gupta2020infotabs,akhtar2022pubhealthtab}, and text generation~\cite{lebret2016neural,wiseman2017challenges,cheng2021hitab}.
Following the success of Multimodal Large Language Models (MLLMs) across domains~\cite{alayrac2022flamingo,li2023blip,liu2023visual,park2024llamo,park2025deepvideo}, MLLM-based architectures~\cite{zhao2024tabpedia,zhou2025syntab,chen2024expanding,zheng2024multimodal} have shown potential in table understanding tasks.
Previous works largely focused on data construction and still struggled to interpret table images. 
Inspired by studies that condition vision encoders on input instructions~\cite{abramovich2024visfocus,ganz2024question}, we enhance table understanding by generating question-aware features while simultaneously reducing output tokens, yielding a more informative and compact representation.

\noindent\textbf{Efficient MLLM.}
MLLMs suffer from high computational costs, primarily due to the quadratic complexity of self-attention in Language Models (LMs) concerning input token length. 
To address this, recent works have explored applying token pruning or merging strategies~\cite{bolya2022token,liang2022not,lee2024multi,choi2024vid} for MLLMs. 
For instance, FastV~\cite{chen2024image}, FitPrune~\cite{ye2025fit}, and SparseVLM~\cite{zhang2024sparsevlm} prune low-attention tokens based on attention scores, while LLaVA-PruMerge~\cite{shang2025prumerge} clusters similar tokens to retain key visual context with fewer tokens. 
These methods either rely on attention scores, which are incompatible with FlashAttention~\cite{dao2022flashattention}, or involve additional similarity computations, adding substantial overhead.
In this work, we propose a simple yet effective pruning method that is fully compatible with FlashAttention and requires negligible extra computation. 
Moreover, unlike prior works that focus solely on better token selection criteria, we introduce a complementary strategy that explicitly encourages the model to retain essential information in the non-pruned tokens, thereby minimizing the information loss induced by pruning.

\section{Method}
In this section, we first outline general MLLM architectures (Sec.~\ref{sec:mllm_overview}). 
We then introduce progressive question conditioning, which injects question embeddings into ViT layers with gradually increasing frequency to produce question-aware visual features (Sec.~\ref{sec:progressive_q_cond}). 
Next, we introduce an $L_2$ norm-based pruning strategy that removes background tokens for efficiency. 
We also propose token focusing, which adapts the model to pruning by encouraging essential information to reside in the retained tokens. 
(Sec.~\ref{sec:token_focusing}).
Finally, we combine these components to form TabFlash, an efficient MLLM for table understanding (Sec.~\ref{sec:tabflash}).

\subsection{Overall architecture of MLLMs}
\begin{figure*}[t!]
\begin{center}
\includegraphics[width=\textwidth]{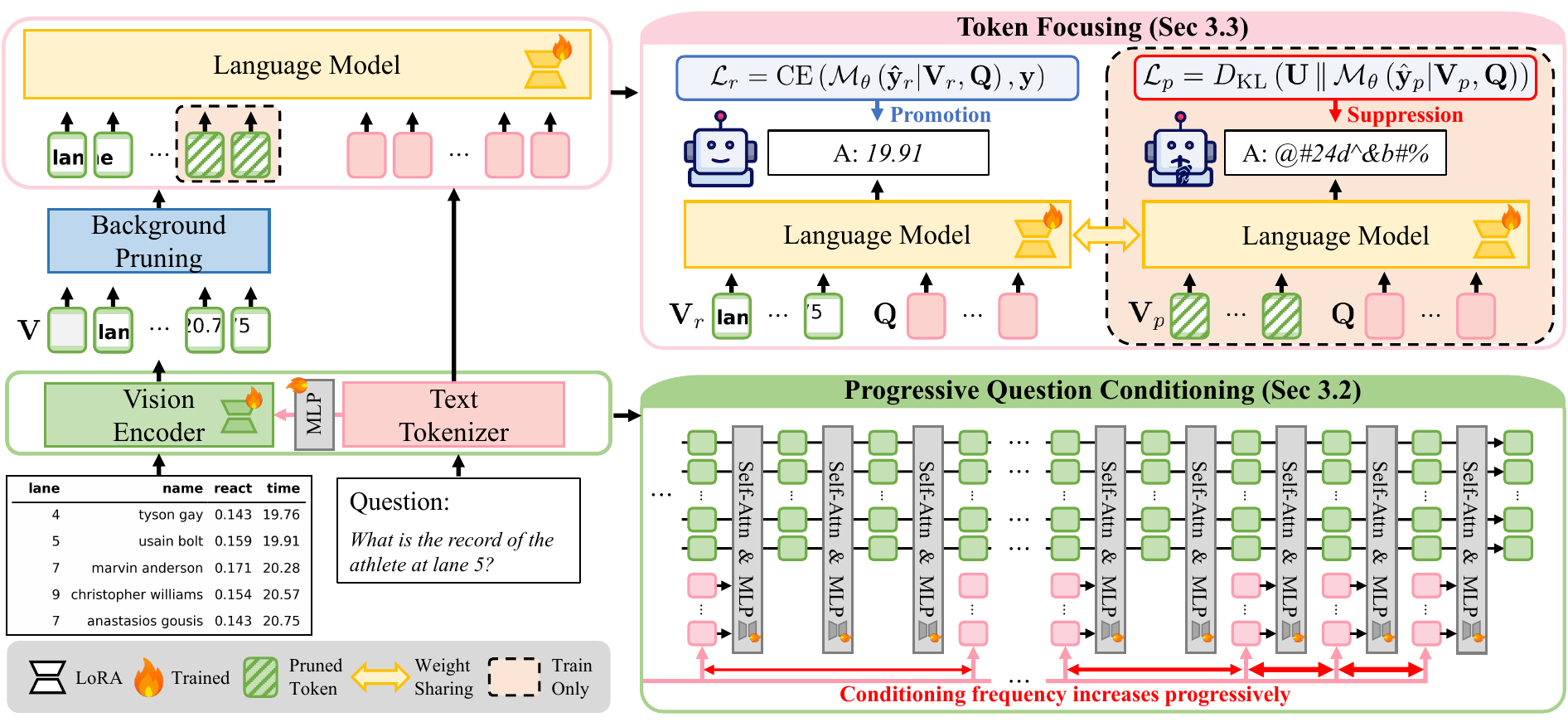}
\end{center}
\caption{
\textbf{Overall pipeline of TabFlash.}
Progressive question conditioning injects question information into ViT layers with a progressively increasing frequency, producing a question-relevant visual token set $\mathbf{V}$ (Sec. \ref{sec:progressive_q_cond}).
The tokens are divided into a pruned set $\mathbf{V}_p$ and a retained set $\mathbf{V}_r$, where only $\mathbf{V}_r$ is used during inference for efficiency.
To concentrate information in $\mathbf{V}_r$, token focusing encourages accurate prediction with $\mathbf{V}_r$ while suppressing prediction using $\mathbf{V}_p$ (Sec. \ref{sec:token_focusing}).
}
\label{fig:main_fig}
\end{figure*}
\label{sec:mllm_overview}
In this section, we outline the general pipeline of Multimodal Large Language Models (MLLMs).
MLLM generates a response given an input image $\mathbf{I}$ and a question $\mathbf{Q}$.
An input image $\mathbf{I}$ is first fed into a Vision Transformer (ViT), which extracts a processed set of visual tokens $\mathbf{V}$, which summarizes the image, through multiple attention layers.
The process within ViT is formally defined as follows.
Provided an input image $\mathbf{I}$, an initial input embedding is generated as $\mathbf{V}_1 = \text{Emb}_v(\mathbf{I}) \in \mathbb{R}^{v \times d}$, where $\text{Emb}_v(\cdot)$ denotes an image embedding layer, $v$ and $d$ denote the number of tokens and ViT feature dimension, respectively.
Then, the embedding is passed through multiple ViT layers with index of $l = 1, 2 \dots, L$, where $L$ denotes the number of ViT layers.
In the $l$-th layer, self-attention is first applied to input $\mathbf{V}_l$, followed by an MLP projection layer, resulting in refined output $\mathbf{V}_{l+1}$, which is fed to the $(l+1)$-th layer as input.
The process is defined as:
\begin{equation}
    \mathbf{V}_{l+1} \in \mathbb{R}^{v\times d} = \text{MLP}_l\left(\text{Self-Attn}_l\left(\mathbf{V}_l\right)\right). \\
\end{equation}
After the $L$-th layer, $\mathbf{V}_{L+1}$ is generated, which is an image representation extracted by ViT.
In the rest of the paper, we denote $\mathbf{V}_{L+1}$ as a visual token set $\mathbf{V}$ for conciseness.
Obtained $\mathbf{V}$ are then fed into a projector (\eg MLP~\cite{liu2023visual}) which maps visual tokens to the language model (LM) space.
Finally, $\mathcal{M}_{\theta}$, an LM parameterized by $\theta$ takes visual tokens $\mathbf{V}$ and a question $\mathbf{Q}$ as input, and generates final response.
The model is trained by minimizing the conventional LLM loss, defined as:
\begin{equation}
    \mathcal{L}_{\text{llm}} = \mathrm{CE}\left(\mathcal{M}_{\theta}\left(\mathbf{\hat{y}}|\mathbf{V}, \mathbf{Q}\right) | \mathbf{y}\right),
    \label{eq:llm_loss}
\end{equation}
where $\mathrm{CE}(\cdot)$ denotes the cross-entropy loss between the prediction $\mathbf{\hat{y}}$ and the correct response $\mathbf{y}$. 
Although the architecture has been successful on natural images, we identify two key limitations when applying it to tabular understanding tasks.
First, the visual token set $\mathbf{V}$ is generated without considering the question $\mathbf{Q}$.
This is especially problematic for table images, where focus on local regions relevant to the question is crucial.
Also, as a large number of visual tokens are generated to represent table images in detail, a significant computational burden is induced, as LM generally has quadratic complexity regarding the number of input tokens.
To this end, we propose a question-conditioning method to generate $\mathbf{V}$ relevant to the question, and a pruning strategy along with tailored fine-tuning objectives to pursue efficiency while minimizing information loss.

\subsection{Progressive Question Conditioning for ViT}
\noindent\textbf{Question conditioning for ViT.}
\label{sec:progressive_q_cond}
To obtain a visual token set $\mathbf{V}$ relevant to the question, we additionally condition ViT layers with the question $\mathbf{Q}$, motivated by previous works~\cite{ganz2024question,abramovich2024visfocus}.
A question $\mathbf{Q}$ is first converted to embeddings with length of $q$ via an embedding function $\text{Emb}_q(\cdot)$, which is a tokenizer of the LM.
Then, a two-layer MLP $\mathcal{P}_l(\cdot)$ projects converted embedding to ViT feature dimension $d$, resulting in question embedding $\mathbf{Q}_l$ for layer $l$ as follows:
\begin{equation}
    \mathbf{Q}_l \in \mathbb{R}^{q\times d} = \mathcal{P}_l\left(\text{Emb}_q\left(\mathbf{Q}\right)\right).
\end{equation}
Then, generated question embedding $\mathbf{Q}_l$ is concatenated to input embedding $\mathbf{V}_l$ as:
\begin{equation}
    \mathbf{V}^c_l \in \mathbb{R}^{(v+q) \times d} = \text{Concat}\left([\mathbf{V}_l, \mathbf{Q}_l]\right),
\end{equation}
forming a combined embedding $\mathbf{V}_l^c$, where $v$ and $q$ denotes number of input and question tokens, respectively.
Then, information between tokens is fused through a self-attention operation as:
\begin{equation}
    \mathbf{V}^{'}_l \in \mathbb{R}^{(v + q) \times d}= \text{Self-Attn}_l\left(\mathbf{V}^c_l\right), \\
\end{equation}
resulting in fused embedding $\mathbf{V}^{'}_l$.
Then, only the first $v$ tokens from $\mathbf{V}^{'}_l$ corresponding to original input tokens are selected and fed into the MLP projection layer as:
\begin{equation}
    \mathbf{V}_{l+1} \in \mathbb{R}^{v\times d} = \text{MLP}_l\left(\mathbf{V}^{'}_{l}[0:v]\right), \\
\end{equation}
resulting in the layer output $\mathbf{V}_{l+1}$ having the same number of tokens as the layer input $\mathbf{V}_l$.
By injecting the question embedding into ViT, the final visual token set $\mathbf{V}$ that is more relevant to the question is obtained.

\noindent\textbf{Progressive question conditioning.}
Although question conditioning might help obtain a better visual representation, selecting layer $l$ to conduct conditioning is non-trivial.
As shown in \cite{ganz2024question}, conditioning on inappropriate layers might even degrade the performance.
To this end, we propose \textit{progressive question conditioning}, which injects the question into ViT layers with progressively increasing frequencies.
In other words, early ViT layers are intermittently conditioned with a large interval, while the interval decreases as the layer progresses, conditioning late layers more frequently.
Such a design choice is grounded on previous observations~\cite{dosovitskiy2020image,raghu2021vision} that early ViT layers are volatile as they focus on details of the image, while latter layers, aggregating global information, are relatively stable.
Intuitively, progressive conditioning allows stable injection of question information by adjusting the conditioning frequency proportional to each layer's capacity to handle the question.
Note that question conditioning adds only a negligible total computation cost of 0.4\%, making it highly efficient.

\subsection{Token Focusing with Background Pruning}
\label{sec:token_focusing}
\noindent\textbf{Background token pruning.}
\begin{figure}[t!]
\begin{center}
\includegraphics[width=0.9\linewidth]{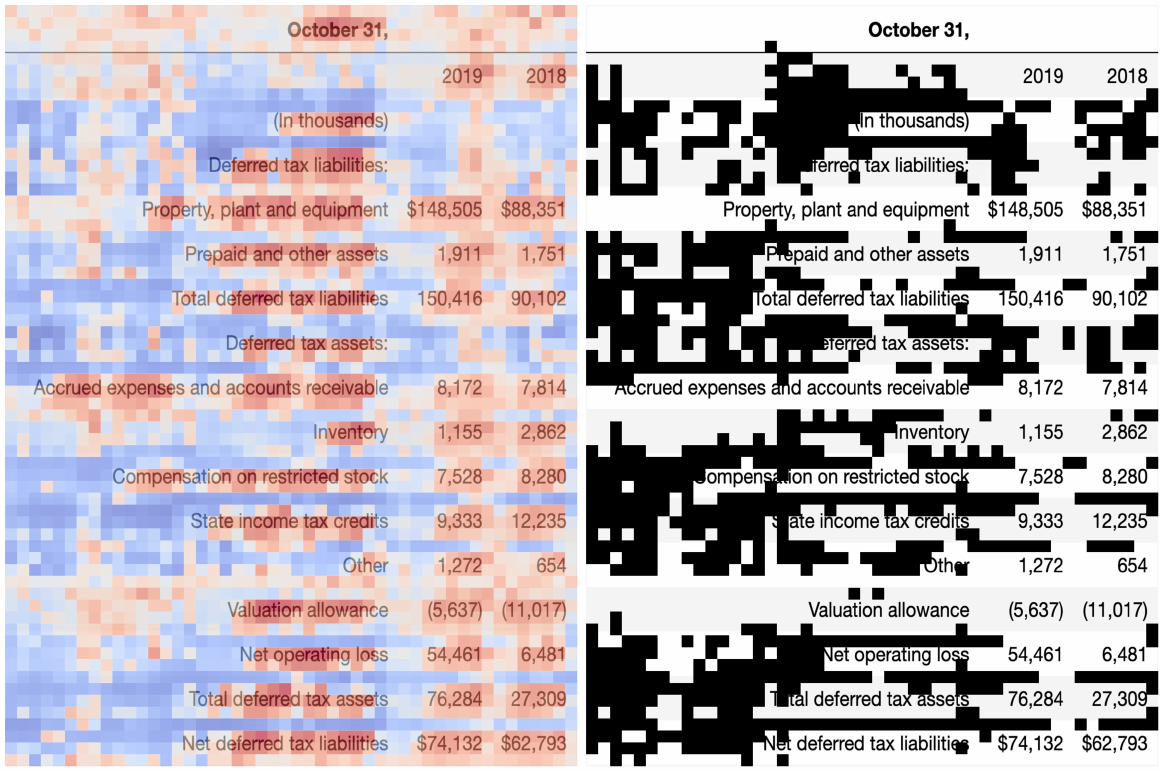}
\end{center}
\caption{
\textbf{Visualization of $L_2$ norms of ViT output tokens (left) and norm-based pruning results (right).}
\textcolor{red}{Red} and \textcolor{blue}{blue} color denotes \textcolor{red}{high} and \textcolor{blue}{low} $L_2$ norms, respectively.
30\% of tokens with the lowest norms are pruned ($p=0.3)$.
More examples provided in the supplementary material.
}
\label{fig:l2_norm_visualization_fig}
\end{figure}
Conventional MLLMs~\cite{chen2024expanding,zhao2024tabpedia} produce up to 2–3k visual tokens, incurring high computational cost due to the quadratic complexity of LMs.
However, tabular images are highly redundant, with a large portion of tokens representing background. 
We observe that the $L_2$ norm of visual tokens effectively distinguishes content from background.
In detail, high norms align with content regions, while low norms indicate background (Fig.~\ref{fig:l2_norm_visualization_fig}, left).
Based on this observation, we propose a background pruning strategy.
Given pruning rate $p$, we take $N_r = \lfloor (1 - p) \cdot v\rfloor$ tokens with the highest norms to get retained set $\mathbf{V}_r$ as:
\begin{equation}
    \mathbf{V}_r = \{\mathbf{v}_i \mid i \in \text{Top-}k\left(\|\mathbf{V}\|_2;N_r\right)\},
\end{equation}
where $\|\mathbf{V}\|_2 = \{\|\mathbf{v}_i\|_2\}_{i=1}^{v}$ is a set consisting of $L_2$ norms of tokens $\mathbf{v}_i \in \mathbf{V}$, and $\text{Top-}k(\cdot;N_r)$ is an operation returning indices of $N_r$ elements with highest values.
The remaining tokens form the pruned token set $\mathbf{V}_p$, formally defined as:
\begin{equation}
    \mathbf{V}_p = \mathbf{V}  \setminus \mathbf{V}_r.
\end{equation}
During inference, only the retained set $\mathbf{V}_r$ is fed to LM.
Visualization of the retained set $\mathbf{V}_r$ (Fig.~\ref{fig:l2_norm_visualization_fig}, right) shows successful removal of background tokens.
Unlike attention-based or similarity-based pruning, our method is compatible with FlashAttention~\cite{dao2022flashattention} and avoids costly similarity map construction, making it highly efficient.

\noindent\textbf{Token focusing.}
While background pruning successfully discards background tokens, simply discarding the pruned set $\mathbf{V}_p$ and using the retained set $\mathbf{V}_r$ for inference leads to a significant performance drop (Tab.~\ref{tab:contrastive_token_sup_ablation_tab}, row 4).
A closer look reveals that the model can still answer questions to some extent using only $\mathbf{V}_p$ (Tab.~\ref{tab:margin_by_pruning_loss_tab}), indicating that important information is still being stored in the pruned tokens.
This behavior is undesirable since only the retained tokens $\mathbf{V}_r$ are used during inference, meaning any useful information in $\mathbf{V}_p$ is lost, aggravating the performance degradation.
To address this, we introduce \textit{token focusing}, a novel training strategy that encourages the model to focus important information in the retained tokens $\mathbf{V}_r$. 
Token focusing explicitly promotes information retention on $\mathbf{V}_r$, while discouraging it on $\mathbf{V}_p$.
Specifically, we define the token promotion loss $\mathcal{L}_r$ to encourage accurate predictions based solely on $\mathbf{V}_r$:
\begin{equation}
    \mathcal{L}_{r} = \mathrm{CE}\left(\mathcal{M}_{\theta}\left(\mathbf{\hat{y}}_r|\mathbf{V}_r, \mathbf{Q}\right), \mathbf{y}\right),
\end{equation}
where $\mathrm{CE}$ denotes the cross-entropy loss, $\mathbf{\hat{y}}_r$ is prediction with $\mathbf{V}_r$ and $\mathbf{y}$ is the target answer.
\begin{table*}[t!]
  \centering
  \begin{adjustbox}{width=\textwidth}
  \begin{tabular}{l|c|ccccccc|c}
    \toprule
    \multirow{2}{*}{Method} & \multirow{2}{*}{LM Size} & \multicolumn{5}{c}{In-domain} & \multicolumn{2}{c|}{Out-domain} & \multirow{2}{*}{Avg.} \\
    \cmidrule(lr){3-7}
    \cmidrule(lr){8-9}
    & & TABMWP & WTQ & HiTab & TAT-QA & FeTaQA & AIT-QA & TabMCQ & \\
    \hline
    \rowcolor{lightgray}
    \multicolumn{10}{l}{\textbf{\textit{Proprietary MLLM}}} \\
    \textcolor{gray}{GPT-4o} & \textcolor{gray}{-} & \textcolor{gray}{47.5} & \textcolor{gray}{25.7} & \textcolor{gray}{10.2} & \textcolor{gray}{25.1} & \textcolor{gray}{10.8} & \textcolor{gray}{18.2} & \textcolor{gray}{40.0} & \textcolor{gray}{25.4} \\
    \textcolor{gray}{Gemini 2.5 Pro~\cite{comanici2025gemini}} & - & \textcolor{gray}{48.1} & \textcolor{gray}{49.2} & \textcolor{gray}{24.2} & \textcolor{gray}{19.6} & \textcolor{gray}{5.4} & \textcolor{gray}{37.6} & \textcolor{gray}{18.2} & \textcolor{gray}{28.9} \\
    \textcolor{gray}{GPT-4V}$^\dagger$ & \textcolor{gray}{-} & \textcolor{gray}{60.0} & \textcolor{gray}{48.0} & \textcolor{gray}{27.5} & \textcolor{gray}{32.5} & \textcolor{gray}{11.0} & \textcolor{gray}{62.5} & \textcolor{gray}{66.0} & \textcolor{gray}{43.9}\\
    \hline
    \rowcolor{lightgray}
    \multicolumn{10}{l}{\textbf{\textit{Open-Source MLLM}}} \\
    LLaVA-1.5~\cite{liu2024improved}         & 7B  &  6.1 &  1.2 & 2.0 & 3.0 &  8.2 & - & - & -\\
    Vary-toy~\cite{wei2024vary}           & 1.8B      &  4.4 &  8.0 & 3.4 & 8.8 &  2.4 & 9.4 & - & -\\
    Monkey~\cite{li2024monkey}             & 7B        & 13.3 &  19.1 & 6.4 & 12.3 &  3.4 & - & 18.9 & -  \\
    TabPedia~\cite{zhao2024tabpedia}    & 7B & 12.3 & 20.4 & 1.2 & 9.7 & 12.5 & 17.2 & 1.0 & 10.6 \\
    mPlug-DocOwl1.5~\cite{hu2024mplug}    & 7B & 11.4 & 26.8 & 11.1 & 12.4 & 3.6 & 46.2 & 3.2 & 16.4 \\
    Table-LLaVA~\cite{zheng2024multimodal}    & 7B  & 57.8 & 18.4 &10.1 & 12.8 & 25.6 & 5.5 & 44.5 & 25.0 \\
    Table-LLaVA~\cite{zheng2024multimodal}    & 13B & 59.8 & 20.4 &10.9 & 15.7 & 28.0 & 6.1 & 51.5 & 27.5\\
    SynTab-LLaVA~\cite{zhou2025syntab}    & 7B & 88.3 & 39.6 & 35.7 & 51.9 & 35.5 & 28.6 & 70.6 & 50.0 \\
    InternVL-2.5~\cite{chen2024expanding}    & 3B & 93.1  & 45.2 & 53.7 & 55.7 & 31.8 & \textbf{58.7} & 64.1 & 57.5 \\
    \rowcolor{lightblue}
    TabFlash &  3B & \textbf{93.7} & \textbf{46.4} & \textbf{60.5} & \textbf{59.9} & \textbf{36.1} & 54.9 & \textbf{71.9} & \textbf{60.5} \\
    \midrule
    \multicolumn{10}{l}{\textbf{\textit{Low-cost models$^{\ddagger}$ ($<$5 TFLOPs)}}} \\
    InternVL-2.5~\cite{chen2024expanding}    & 1B & 86.3 & 30.5 & 32.3 & 35.5 & 26.6 & 39.8 & 50.0 & 43.0 \\
    \rowcolor{lightblue}
    TabFlash &  1B & 88.5 & 32.1 & 40.9 & 44.8 & 32.9 & 41.9 & 57.8 & 
    48.4\\
    \bottomrule
  \end{tabular}
  \end{adjustbox}
  \caption{\textbf{Results on table question-answering benchmarks.}
  Best results among open-source MLLMs highlighted \textbf{bold}.
  Avg.: performance averaged over seven benchmarks.
  Out-domain: model is not trained with the training set of the dataset.
  $\dagger$: evaluation results on subset~\cite{zheng2024multimodal}.
  $\ddagger$: models with exceptionally low cost ($<$ 5 TFLOPs).
  See Tab.~\ref{tab:cost_comparison} for cost analysis.
  }
  \label{tab:main_table}
\end{table*}
On the other hand, to suppress the information retention in $\mathbf{V}_p$, we define a token suppression loss $\mathcal{L}_p$ as the KL divergence between the model’s prediction based on $\mathbf{V}_p$ and a uniform distribution $\mathbf{U}$ over the vocabulary:
\begin{equation}
    \mathcal{L}_{p} = D_{\mathrm{KL}}\left(\mathbf{U} \,\|\, \mathcal{M}_{\theta}\left(\mathbf{\hat{y}}_p|\mathbf{V}_p, \mathbf{Q}\right)\right),
    \label{eq:suppression_loss}
\end{equation}
where $\mathbf{\hat{y}}_p$ is prediction with $\mathbf{V}_p$.
Minimizing $\mathcal{L}_p$ penalizes meaningful prediction from $\mathbf{V}_p$, thereby discouraging the storage of meaningful information in pruned tokens.
Combining both objectives, the token focusing loss is defined as:
\begin{equation}
    \mathcal{L}_{\text{tf}} = \mathcal{L}_{r} + \lambda \cdot \mathcal{L}_{p},
    \label{eq:contrastive_token_supervision}
    \end{equation}
where $\lambda$ is a hyperparameter.
Unlike existing pruning methods that aim to find better $\mathbf{V}_p$ with minimal information loss, token focusing takes a complementary approach: it regularizes the location of information storage itself, guiding the model to preserve critical content within $\mathbf{V}_r$ rather than $\mathbf{V}_p$.
Our analysis (Sec.~\ref{sec:token_focusing_analysis}) demonstrates that token focusing contributes to the transfer of information from $\mathbf{V}_p$ to $\mathbf{V}_r$, thereby minimizing the information loss by pruning.

\subsection{TabFlash}
\label{sec:tabflash}
\noindent\textbf{TabFlash.}
Combining progressive question conditioning and pruning with token focusing, we present TabFlash.
Equipped with both methods, a compact and informative token set $\mathbf{V}_r$ is obtained, improving both the effectiveness and efficiency.
Fig.~\ref{fig:main_fig} outlines the overall pipeline of TabFlash.

\noindent\textbf{Training process.}
We train TabFlash in two stages. 
First, the model is trained for 3 epochs using progressive question conditioning with conventional LLM loss $\mathcal{L}_{\text{llm}}$ (Eq.~\eqref{eq:llm_loss}). 
Then, it is trained for an additional epoch with both progressive question conditioning and background pruning applied, utilizing the token focusing loss $\mathcal{L}_{\text{tf}}$ (Eq.~\eqref{eq:contrastive_token_supervision}). 
The 3B variant model is efficiently trained within 22 hours using 8 NVIDIA H200 GPUs.

\section{Experiments}
\subsection{Implementation Details.}
We develop TabFlash by fine-tuning the 1B and 3B variants of InternVL-2.5~\cite{chen2024expanding}.
Following Table-LLaVA~\cite{zheng2024multimodal}, we train on the MMTab-pre and MMTab-instruct datasets, totaling 383k samples.
LoRA~\cite{hu2022lora} with rank 16 is applied to both the ViT and LLM, while all other parameters remain frozen.
Training is conducted for 4 epochs with a learning rate of $4e{-5}$, $\lambda = 2e{-4}$, and pruning rate $p = 0.3$.
The exact configuration of question conditioning layers is in the supplementary material.

\subsection{Datasets and Evaluation Metrics.}
We evaluate the performance on seven table question-answering datasets, TABMWP~\cite{lu2022dynamic}, WTQ~\cite{pasupat2015compositional}, HiTab~\cite{cheng2021hitab}, TAT-QA~\cite{zhu2021tat}, FeTaQA~\cite{nan2022fetaqa}, AIT-QA~\cite{katsis2021ait}, and TabMCQ~\cite{jauhar2016tabmcq} over 17.9k samples in total.
For evaluation protocol and metrics, we directly follow the settings of Table-LLaVA~\cite{zheng2024multimodal} without any modification. 
Accuracy is adopted as a metric for all datasets except for FeTaQA, adopting BLEU~\cite{papineni2002bleu} as a metric.

\subsection{Main Results.}
In Tab.~\ref{tab:main_table}, we compare TabFlash with proprietary and open-source MLLMs on seven table QA benchmarks.
While QA is our main focus, we also report results on fact verification and table-to-text generation in the supplementary material.

\noindent\textbf{Results.}
TabFlash (3B) achieves the highest average performance (60.5), outperforming all open-source models on six of seven benchmarks and ranking second on AIT-QA.
The 1B variant also performs well, surpassing most prior models with an average score of 48.4.
Importantly, this is achieved with significantly lower computational cost, as detailed in the next section.
Notably, TabFlash also outperforms proprietary models (\eg GPT-4V, Gemini 2.5 Pro) on average, highlighting both the difficulty of table understanding and the effectiveness of our approach.

\begin{table}[t!]
    \centering
    \begin{adjustbox}{width=\linewidth}
    \begin{tabular}{l|c|cc|c}
    \toprule
      Method & LM Size &TFLOPs $\downarrow$& Memory$\downarrow$ & Avg.$\uparrow$ \\
    \midrule
    TabPedia & 7B& 27.08 & 22.7G & 10.6 \\ 
    DocOwl1.5 & 7B & 26.60 & 24.0G & 16.4 \\
    Table-LLaVA & 7B & 10.42 & 15.0G & 25.0 \\  
    Table-LLaVA & 13B & 18.47 & 28.1G & 27.5 \\  
    SynTab-LLaVA & 7B & 15.21 & 16.4G & 50.0 \\ 
    InternVL-2.5 & 3B & 14.23 & 24.7G & 57.5 \\ 
    \rowcolor{lightblue}
    TabFlash & 3B & 10.38 & 17.3G & \textbf{60.5} \\ 
    \midrule
    \multicolumn{5}{l}{\textbf{\textit{Low-cost models ($<$5 TFLOPs)}}} \\
    InternVL-2.5  & 1B & 2.59 & 18.5G & 43.0 \\ 
    \rowcolor{lightblue}
    TabFlash & 1B & \textbf{1.78}& \textbf{11.2G} & 48.4 \\ 
    \bottomrule 
    \end{tabular}
    \end{adjustbox}
    \caption{\textbf{Cost Analysis.}
    Memory: Peak GPU memory usage.
    }
    \label{tab:cost_comparison}
\end{table}
\subsection{Cost Analysis.}
\label{sec:cost_analysis}
In Tab.~\ref{tab:cost_comparison}, we report the LLM TFLOPs and peak GPU memory usage of TabFlash and previous open-source MLLMs.
TabFlash (3B) achieves the best overall performance while maintaining high efficiency, requiring only 10.38 TFLOPs and 17.3 GB of memory.
Compared to the second-best model, InternVL-2.5 (3B), TabFlash reduces FLOPs by 27\% and memory usage by 30\%, while achieving a 3-point gain in accuracy.
The 1B variant of TabFlash further improves computational efficiency, consuming just 1.78 TFLOPs and 11.2 GB of memory, and still outperforms five of the seven baselines.
Compared to SynTab-LLaVA with a difference of 1.6\%p in accuracy, TabFlash (1B) requires 88\% fewer FLOPs and 32\% less memory, highlighting its exceptional cost-effectiveness for resource-constrained scenarios.
Latency analysis is provided in the supplementary material.

\section{Analysis}
In this section, we present various analyses to investigate the effect of each component. 
All experiments report the average accuracy achieved by TabFlash (1B). 
Detailed results for each dataset are provided in the supplementary material.
\begin{table}[t!]
    \centering
    \begin{adjustbox}{width=\linewidth}
    \begin{tabular}{l|ccc}
    \toprule
    Injection layers & Interval & \# cond. layers & Avg. \\
    \midrule
    All (1-24) & 1 & 24 & 43.4 \\
    Early (1-8) & 1 & 8 & 44.8 \\
    Mid (9-16)& 1 & 8 & 48.0 \\
    Late (17-24)& 1  & 8 & 48.8 \\
    \midrule
    \multirow{3}{*}{Sparse} & 2 & 12 & 48.0  \\
     & 3 & 8 & 48.6  \\
     & 4 & 6 & 49.0 \\
    \midrule
    \rowcolor{lightblue}
    Progressive (Ours)& Progressive & 6 & \textbf{50.3} \\
    \midrule
    \end{tabular}
    \end{adjustbox}
    \caption{\textbf{Ablation on progressive question conditioning.}
    \# cond. layers: total \# layers conditioning is applied to.
    }
    \label{tab:progressive_cond_ablation_tab}
\end{table}
\begin{table}[t!]
    \centering
    \begin{adjustbox}{width=\linewidth}
    \begin{tabular}{l|c|c|c}
    \toprule
      Method & TFLOPs $\downarrow$ & Pruning & Avg.$\uparrow$ \\
    \midrule
    Upper bound (unpruned) &2.59 & & 50.3 \small \color{ForestGreen}{(100\%)} \\ 
    \midrule
    \rowcolor{lightblue}
    Ours (both $\mathcal{L}_p$, $\mathcal{L}_r$) & 1.78&\checkmark & 48.4 \small \color{ForestGreen}{(96\%)}\\
    \ \ - without $\mathcal{L}_p$ & 1.78&  \checkmark & 47.3 \small \color{ForestGreen}{(94\%)}\\
    \ \ \ \ \ - without $\mathcal{L}_r$ & 1.78& \checkmark & 37.2 \small \color{ForestGreen}{(74\%)} \\
    \midrule
    \end{tabular}
    \end{adjustbox}
    \caption{\textbf{Ablation on token focusing loss.}
    $\checkmark$ indicates pruning applied for inference. 
    Percentages are relative to the unpruned baseline (first row).
    }
    \label{tab:contrastive_token_sup_ablation_tab}
\end{table}
\subsection{Ablation Study.}
\noindent\textbf{Progressive question conditioning.}
In Tab.~\ref{tab:progressive_cond_ablation_tab}, we compare conditioning strategies based on injection layer choices. 
We evaluate conditioning at all layers (\q{All}), as well as selectively at early (\q{Early}), mid (\q{Mid}), and late (\q{Late}) layers, all with a fixed interval of 1. 
We then consider sparsely conditioning at every 2, 3, or 4 layers (\q{Sparse}). 
Finally, we propose conditioning with a progressively decreasing interval, where the conditioning frequency increases as the layer progresses (\q{Progressive}).
Please refer to the supplementary material for more detailed configurations.
Results suggest that early layers are particularly sensitive to conditioning, and overly frequent injection degrades performance, making stable information integration difficult.
Nevertheless, our method achieves the best performance, demonstrating its robustness in injecting question information across layers.

\noindent\textbf{Token focusing.}
In Tab.~\ref{tab:contrastive_token_sup_ablation_tab}, we present ablation results for token focusing. 
Compared to TabFlash without pruning, applying pruning with the proposed token focusing reduces FLOPs by 31\%, while limiting the performance degradation to only 3.8\%. 
Removing the token suppression loss ($\mathcal{L}_p$) leads to an additional performance drop of 1.1 points, showing its importance. 
Furthermore, removing the token promotion loss ($\mathcal{L}_r$) results in a substantial decline in performance. 
These results highlight the significance of token focusing, demonstrating the necessity of a proper fine-tuning strategy facilitating the model's adaptation to pruning results by enforcing information concentration on the retained token set.

\begin{table}[t!]
    \centering
    \begin{adjustbox}{width=\linewidth}
    \begin{tabular}{c|c|cc}
    \toprule
      Training Loss & Token set for inference & Avg. & $\Delta$ \\
    \midrule
    \multirow{2}{*}{$\mathcal{L}_r$} & $\mathbf{V}_r$ & 47.3 & \multirow{2}{*}{33.8} \\
     & $\mathbf{V}_p$ & 13.5 & \\
    \midrule
    \multirow{2}{*}{$\mathcal{L}_r, \mathcal{L}_p$} & $\mathbf{V}_r$ & 48.4 & \multirow{2}{*}{48.4} \\
     & $\mathbf{V}_p$ & 0.0 &  \\
    \bottomrule
    \end{tabular}
    \end{adjustbox}
    \caption{\textbf{Performance comparison using $\mathbf{V}_r$ and $\mathbf{V}_p$ for inference.}
    $\Delta$ denotes the difference in average performance between the retained set $\mathbf{V}_r$ and pruned set $\mathbf{V}_p$ for inference.
    }
    \label{tab:margin_by_pruning_loss_tab}
\end{table}
\begin{figure*}[t!]
\begin{center}
\includegraphics[width=1.0\textwidth]{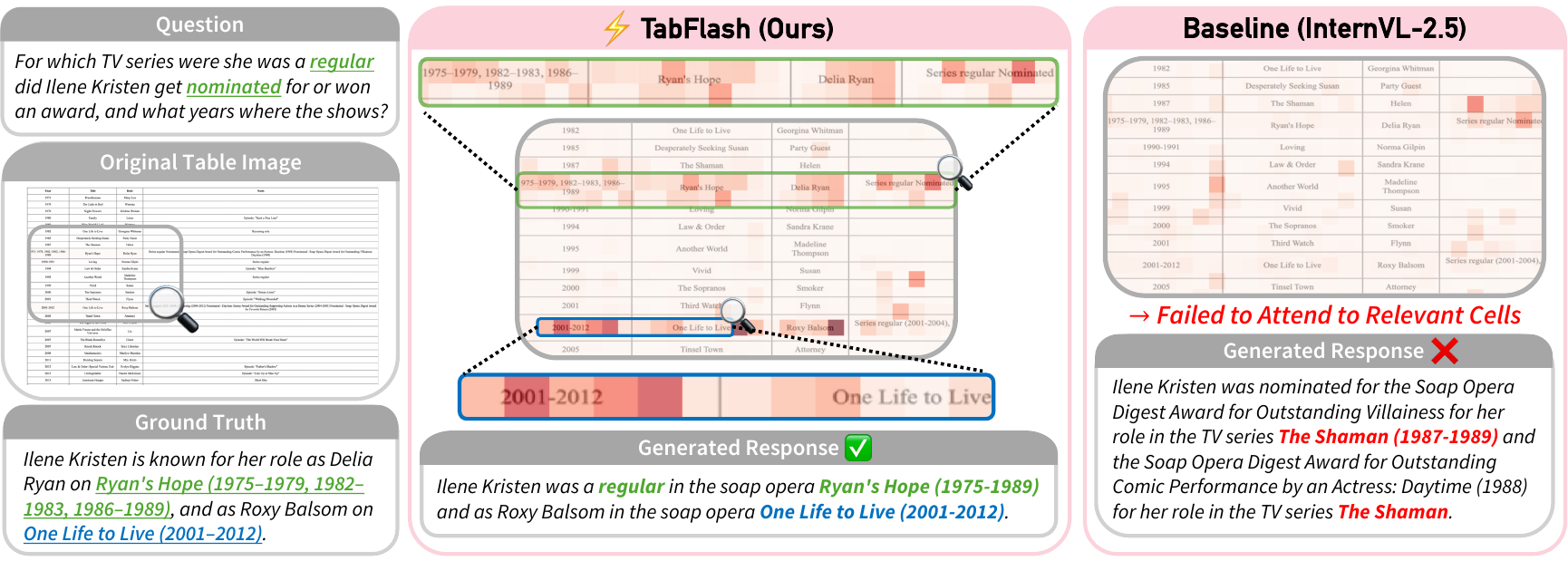}
\end{center}
\caption{
\textbf{Qualitative results.}
Near-white regions indicate low attention, while stronger red colors represent higher attention scores.
Best viewed when zoomed in.
Please refer to the supplementary material for further qualitative results.
}
\label{fig:qualitative_fig}
\end{figure*}
\subsection{Effect of Token Focusing.}
\label{sec:token_focusing_analysis}
To investigate whether the suppression loss operates as desired, we compare the performance of models trained with and without the suppression loss $\mathcal{L}_p$, evaluating when different token sets are provided to the LM for inference (Tab.~\ref{tab:margin_by_pruning_loss_tab}).
When trained solely with the token promotion loss $\mathcal{L}_r$, accuracy of 13.5 is achieved even using the pruned set $\mathbf{V}_p$, indicating the presence of useful information in $\mathbf{V}_p$. 
Adding the suppression loss $\mathcal{L}_p$ addresses this by penalizing the model’s ability to answer correctly from $\mathbf{V}_p$, thereby concentrating valuable information in the retained set $\mathbf{V}_r$.
Thus, the model trained with $\mathcal{L}_p$ shows 1.1 points higher accuracy with $\mathbf{V}_r$ than without $\mathcal{L}_p$.
Also, the gap between $\mathbf{V}_r$ and $\mathbf{V}_p$ widens from 33.8 to 48.4 points, indicating information transfer from $\mathbf{V}_p$ to $\mathbf{V}_r$.
These results validate the effectiveness of token focusing for retaining information on $\mathbf{V}_r$, making the model more compatible with the pruning strategy.

\subsection{Analysis on Pruning Rate $p$.}
\begin{table}[t!]
    \centering
    \begin{adjustbox}{width=\linewidth}
    \begin{tabular}{c|cc|c}
    \toprule
      Pruning rate $p$ & TFLOPs $\downarrow$& Memory$\downarrow$ & Avg.$\uparrow$ \\
    \midrule
    0.0 & 2.59 \small \color{ForestGreen}{(100\%)}& 18.7G \small \color{ForestGreen}{(100\%)} & 50.3 \small \color{ForestGreen}{(100\%)} \\ 
    \midrule
    0.1 & 2.31 \small\color{ForestGreen}{(89\%)}& 15.3G \small \color{ForestGreen}{(82\%)} & 49.4 \small \color{ForestGreen}{(98\%)} \\  
    0.2 & 2.03 \small\color{ForestGreen}{(78\%)}& 12.4G \small \color{ForestGreen}{(66\%)} & 48.5 \small \color{ForestGreen}{(96\%)} \\ 
    \rowcolor{lightblue}
    0.3 & 1.78 \small\color{ForestGreen}{(69\%)}& 11.2G \small \color{ForestGreen}{(60\%)} & 48.4 \small \color{ForestGreen}{(96\%)} \\ 
    0.4 & 1.52 \small\color{ForestGreen}{(59\%)}& 11.1G \small \color{ForestGreen}{(59\%)} & 46.7 \small \color{ForestGreen}{(93\%)} \\ 
    0.5 & 1.28 \small\color{ForestGreen}{(49\%)}& 11.0G \small \color{ForestGreen}{(59\%)} & 44.0 \small \color{ForestGreen}{(87\%)} \\ 
    \bottomrule 
    \end{tabular}
    \end{adjustbox}
    \caption{\textbf{Analysis on pruning rate $p$.}
    Percentages are relative to the unpruned baseline (first row).
    }
    \label{tab:pruning_rate_analysis}
\end{table}
Tab.~\ref{tab:pruning_rate_analysis} reports TFLOPs, peak memory, and accuracy across pruning rates ($p$).
Without pruning ($p=0.0$), the model achieves its full performance of 50.3. 
At $p=0.1$, the model retains 98.2\% of its original performance, while reducing TFLOPs and memory usage by 11\% and 18\%, respectively.
A more aggressive pruning rate of $p=0.5$ yields substantial efficiency gains, cutting TFLOPs by 51\% and memory by 41\%.
These results show that the pruning rate offers a tunable trade-off between performance and efficiency.
We set the default rate to $p=0.3$, achieving 31\% TFLOPs and 40\% memory reduction with only a modest performance drop.

\subsection{Performance Comparison by Table Size.}
\begin{table}[t!]
    \centering
    \begin{adjustbox}{width=\linewidth}
    \begin{tabular}{c|ccc}
    \toprule
      Method & Small & Medium & Large \\
    \midrule
    InternVL-2.5 (1B) & 85.1 & 64.1 & 29.9 \small \\
    \rowcolor{lightblue}
    TabFlash (1B) & 87.6 \small \color{ForestGreen}{(+3\%)}& 67.6 \small \color{ForestGreen}{(+5\%)} & 34.4 \small \color{ForestGreen}{(+15\%)}\\
    \bottomrule 
    \end{tabular}
    \end{adjustbox}
    \caption{
    \textbf{Performance by table size.} 
    Small, Medium, and Large each include equal-sized subsets sorted by image size.
    }
    \label{tab:performance_by_table_size}
\end{table}
In Tab.~\ref{tab:performance_by_table_size}, we compare the accuracy of TabFlash with the baseline architecture, InternVL, across three subsets divided by image sizes: small, medium, and large.
Difficulty increases proportionally to the image size, as the portion of the question-relevant region gets smaller in larger images.
The performance gap between baseline and TabFlash widens as the size of the image enlarges, reaching the 15\% gap in the \q{large} group.
These results highlight TabFlash's capability in capturing fine-grained details in larger images with informative visual tokens.

\subsection{Qualitative Analysis.}
Fig.~\ref{fig:qualitative_fig} illustrates the average attention each visual token receives during generation, where higher scores indicate greater information extraction from the token.
TabFlash focuses attention on question-relevant rows that occupy only a small part of the image, while the baseline distributes attention broadly, overlooking key regions.
As a result, TabFlash correctly generates the answer (\eg \dq{Ryan’s Hope}, \dq{One Life to Live}, and the corresponding years), accurately reflecting the question’s intent.
In contrast, the baseline produces an incorrect response (\eg \dq{The Shaman}), describing entries from irrelevant rows due to its failure to focus on the question-relevant area.
These results show that progressive question conditioning enables question-specific visual features, leading to more focused and accurate attention.

\section{Conclusion}
We present TabFlash, an efficient MLLM for table understanding.
Progressive question conditioning injects the question into ViT layers at progressively increasing frequency.
Background pruning discards redundant tokens based on the norm of each token, effectively removing uninformative regions.
Token focusing minimizes the information loss induced by pruning by enforcing the concentration of essential information on retained tokens.
Together, TabFlash achieves the state-of-the-art performance with significantly reduced computational cost.

\section*{Acknowledgments}
This work was partly supported by Institute for Information \& communications Technology Promotion (IITP) grant funded by the Korea government (MSIP) (No. RS-2024-00443251, Accurate and Safe Multimodal, Multilingual Personalized AI Tutors, 40\%), Institute of Information \& communications Technology Planning \& Evaluation (IITP) grant funded by the Korean government (MSIT) (No. RS-2024-00457882, AI Research Hub Project, 30\%), and National Supercomputing Center with supercomputing resources including technical support (KSC-2025-CRE-0085, 30\%).

\bibliography{aaai2026}
\clearpage

\twocolumn[{
\begin{center}
\vspace*{1em}
{\LARGE \bfseries Supplementary Material of\par}
\vspace{6pt}
{\Large TabFlash: Efficient Table Understanding with Progressive Question Conditioning and Token Focusing\par}
\vspace{1em}
\end{center}
}]

\setcounter{section}{0}
\setcounter{table}{0}
\setcounter{figure}{0}
\renewcommand{\thesection}{\Alph{section}}
\renewcommand{\thetable}{\Alph{section}\arabic{table}}
\renewcommand{\thefigure}{\Alph{section}\arabic{figure}}
The supplementary material is organized into the following sections.
\begin{outline}
    \1 Section \ref{app_sec:add}: Additional Experimental Results
    \2 Section \ref{app_sec:ablation}: Full Results for Ablation Studies
    \2 Section \ref{app_sec:other}: Results on Other Tabular Tasks
    \2 Section \ref{app_sec:suppression_loss_analysis}: Analysis on Suppression Loss Choice
    \2 Section \ref{app_sec:latency_analysis}: Latency Analysis
    \1 Section \ref{app_sec:detailed}: Detailed Experimental Settings
    \2 Section \ref{app_sec:proprietary_mllms}: Proprietary MLLMs
    \2 Section \ref{app_sec:question_cond_configurations}: Detailed Configurations about Question Conditioning
    \1 Section \ref{app_sec:l2norm}: More Visualizations of ViT $L_2$ norms
    \1 Section \ref{app_sec:qual}: More Qualitative Results
\end{outline}

\section{Additional Experimental Results}
\label{app_sec:add}
\subsection{Full Results for Ablation Studies}
\label{app_sec:ablation}
Here, we provide detailed results for each dataset examined in the ablation studies in Sec.~5.
In detail, full results of Tab.~3, Tab.~4, Tab.~5, and Tab.~6 in the main paper are reported on Tab.~\ref{tab:progressive_cond_ablation_full_tab}, Tab.~\ref{tab:contrastive_token_sup_ablation_full_tab}, Tab.~\ref{tab:margin_by_pruning_loss_full_tab}, and Tab.~\ref{tab:pruning_rate_analysis_full}, respectively. 

\subsection{Results on Other Tabular Tasks}
\label{app_sec:other}
In Tab.~\ref{tab:table_tfv_t2t}, we present the performance of TabFlash and baseline models on table fact verification and table-to-text generation tasks.
For fact verification, we evaluate on TabFact~\cite{chen2019tabfact}, InfoTabs~\cite{gupta2020infotabs}, and PubHealthTab~\cite{akhtar2022pubhealthtab} benchmarks.
For table-to-text generation, we evaluate on HiTab\_t2t~\cite{cheng2021hitab}, Rotowire~\cite{wiseman2017challenges}, and WikiBio~\cite{lebret2016neural} benchmarks.
TabFlash achieves the best results on all fact verification benchmarks, outperforming the second-best open-source MLLM by 6.6, 5.5, and 4.9 points on TabFact, InfoTabs, and PubHealthTab, respectively.
It also performs strongly in text generation, achieving the best result on HiTab and competitive scores on Rotowire and WikiBio.
Notably, TabFlash outperforms even GPT-4V while maintaining significantly lower computational cost compared to other MLLMs, demonstrating its efficiency and effectiveness across diverse table understanding tasks.
\begin{table}[t!]
    \centering
    \begin{adjustbox}{width=\linewidth}
    \begin{tabular}{l|c|c|c}
    \toprule
      Method & LM Size & Throughput $\uparrow$ & Avg.$\uparrow$ \\
    \midrule
    TabPedia & 7B & - & 10.6 \\ 
    DocOwl1.5 & 7B & 9.9 & 16.4 \\
    Table-LLaVA & 7B & 37.9 & 25.0 \\  
    Table-LLaVA & 13B & 24.6 & 27.5 \\  
    SynTab-LLaVA & 7B & 31.4 & 50.0 \\ 
    InternVL-2.5 & 3B & 24.9 & 57.5 \\ 
    \rowcolor{lightblue}
    TabFlash & 3B & 27.7 & 60.5 \\ 
    \midrule
    \multicolumn{4}{l}{\textbf{\textit{Low-cost models ($<$5 TFLOPs)}}} \\
    InternVL-2.5  & 1B & 36.1 & 43.0 \\ 
    \rowcolor{lightblue}
    TabFlash & 1B & 39.7 & 48.4 \\ 
    \bottomrule 
    \end{tabular}
    \end{adjustbox}
    \caption{\textbf{Latency Analysis.} Throughput: \# tokens generated per second.
    }
    \label{tab:latency_comparison}
\end{table}
\begin{table*}[t!]
  \centering
  \begin{adjustbox}{width=\textwidth}
  \begin{tabular}{l|c|ccc|ccc|cc}
    \toprule
    \multirow{2}{*}{Method} & \multirow{2}{*}{LM Size} & \multicolumn{3}{c|}{Table Fact Verification} & \multicolumn{3}{c|}{Table-to-Text Generation} & \multirow{2}{*}{TFLOPs} & \multirow{2}{*}{Memory} \\
    \cmidrule(lr){3-8}
    & & TabFact & InfoTabs & PubHealthTab$^*$ & HiTab\_t2t & Rotowire & WikiBio & \\
    \hline
    \rowcolor{lightgray}
    \multicolumn{10}{l}{\textbf{\textit{Proprietary MLLM}}} \\
    \textcolor{gray}{GPT-4V}$^{\dagger}$ & \textcolor{gray}{-} & \textcolor{gray}{45.5} & \textcolor{gray}{65.6} & \textcolor{gray}{67.0} & \textcolor{gray}{3.0} & \textcolor{gray}{4.2} & \textcolor{gray}{1.9} & \textcolor{gray}{-}\\
    \hline
    \rowcolor{lightgray}
    \multicolumn{10}{l}{\textbf{\textit{Open-Source MLLM}}} \\
    BLIP2~\cite{li2023blip}              & 3B     &  18.6 & 27.5 & - & 2.6 & 1.1 & 0.7 & - & - \\
    MiniGPT-4~\cite{zhu2023minigpt}          & 7B      &  0.0 &  0.1 & - & 0.1 & 1.3 & 0.3 & - & -\\
    Qwen-VL~\cite{Qwen-VL}            & 7B        &  1.1 &  0.7 & - & 0.2 & 0.0 & 0.0 & -& -\\
    InternLM-XComposer~\cite{zhang2023internlm} & 7B    &  1.2 &  1.1 & - & 3.3 & 0.4 & 1.5 & -& -\\
    mPLUG-Owl2~\cite{ye2024mplug}         & 7B     & 8.2 & 26.2 & - & 2.1 & 1.2 & 2.2 & -& -\\
    LLaVA-1.5~\cite{liu2024improved}         & 7B  &  18.9 & 28.3 & - & 2.1 & 1.9 & 2.3 & -& -\\
    Vary-toy~\cite{wei2024vary}           & 1.8B      & 6.3 & 7.0 & -  & 0.3 & 0.5 & 0.4 & -& -\\
    Monkey~\cite{li2024monkey}             & 7B        & 22.6 & 22.1 & -  & 1.1 & 0.0 & 2.8 & - & - \\
    TabPedia~\cite{zhao2024tabpedia}    & 7B & 28.8 & 9.2 & 21.0  & 1.2 & 0.0 & 1.1 & 27.08 & 22.7G \\
    mPlug-DocOwl1.5~\cite{hu2024mplug}    & 7B & 27.7 & 28.7 & 28.4 & 3.8 & 0.0 & 0.0 & 26.60 & 24.0G\\
    Table-LLaVA~\cite{zheng2024multimodal}    & 7B  & 59.9 & 65.3 & 51.0  & 9.7 & 10.5 & 9.7 & 10.42 & 15.0G \\
    Table-LLaVA~\cite{zheng2024multimodal}    & 13B & 65.0 & 66.9 & 48.5  & 10.4 & 8.8 & 9.7 & 18.47 & 28.1G\\
    SynTab-LLaVA~\cite{zhou2025syntab}    & 7B & 70.8 & 69.4 & 68.0  & 14.2 & \textbf{14.1} & \textbf{14.1} & 15.21 & 16.4G\\
    InternVL-2.5~\cite{chen2024expanding}    & 3B & 66.0 & 65.3 & 64.4 & 14.5 & 4.5 & 13.0 & 14.23 & 24.7G\\
    \rowcolor{lightblue}
    TabFlash &  3B & \textbf{77.4} & \textbf{74.9} & \textbf{72.9} & \textbf{18.7} & 10.0 & 10.6 & 10.38 & 17.3G\\
    \midrule
    \multicolumn{9}{l}{\textbf{\textit{Low-cost models$^{\ddagger}$ ($<$5 TFLOPs)}}} \\
    InternVL-2.5~\cite{chen2024expanding}    & 1B & 50.6 & 38.0 & 48.8 & 9.4  & 2.9 & 9.2 & 2.59 & 18.5G\\
    \rowcolor{lightblue}
    TabFlash &  1B & 57.7 & 59.7 & 56.2 & 15.3 & 4.1 & 9.3 & 1.78 & 11.2G \\

    \bottomrule
  \end{tabular}
  \end{adjustbox}
  \caption{\textbf{Results on table fact verification and table-to-text generation benchmarks.}
  Best results among open-source MLLMs are highlighted \textbf{bold}.
  $*$: out-domain dataset.
  $\dagger$: evaluation results on subset~\cite{zheng2024multimodal}.
  $\ddagger$: models with exceptionally low cost ($<$ 5 TFLOPs).
  }
  \label{tab:table_tfv_t2t}
\end{table*}
\begin{table*}[t!]
    \centering
    \begin{adjustbox}{width=\textwidth}
    \begin{tabular}{l|cc|ccccccc|c}
    \toprule
    Injection layers & Interval & \# cond. layers & TABMWP & WTQ & HiTab & TAT-QA & FeTaQA & AIT-QA & TabMCQ & Avg. \\
    \midrule
    All (1-24) & 1 & 24 & 66.5 & 32.3 & 40.8 & 39.0 & 32.2 & 38.4 & 54.3 & 43.4 \\
    Early (1-8) & 1 & 8 & 71.6 & 32.1 & 44.7 & 42.0 & 32.5 & 36.8 & 53.7 & 44.8 \\
    Mid (9-16)& 1 & 8 & 83.5 & 32.3 & 42.5 & 45.2 & 33.0 & 41.5 & 57.7 & 48.0 \\
    Late (17-24)& 1  & 8 & 88.7 & 30.2 & 42.1 & 46.1 & 30.2 & 41.7 & 62.0 & 48.8 \\
    \midrule
    \multirow{3}{*}{Sparse} & 2 & 12 & 86.2 & 33.3 & 45.4 & 44.6 & 32.8 & 37.2 & 56.2 & 48.0 \\
     & 3 & 8 & 88.2 & 33.2 & 46.3 & 45.5 & 33.3 & 40.3 & 53.4 & 48.6  \\
     & 4 & 6 & 88.5 & 34.1 & 47.3 & 46.4 & 32.7 & 38.2 & 55.8 & 49.0  \\
    \midrule
    \rowcolor{lightblue}
    Progressive (Ours)& Progressive & 6 & 88.9 & 33.1 & 45.1 & 46.8 & 33.0 & 43.6 & 61.3 & 50.3 \\
    \midrule
    \end{tabular}
    \end{adjustbox}
    \caption{\textbf{Ablation on progressive question conditioning.}
    \# cond. layers: total \# layers conditioning is applied to.
    }
    \label{tab:progressive_cond_ablation_full_tab}
\end{table*}
\begin{table*}[t!]
    \centering
    \begin{adjustbox}{width=\textwidth}
    \begin{tabular}{l|c|c|ccccccc|c}
    \toprule
      Method & TFLOPs $\downarrow$ & Pruning & TABMWP & WTQ & HiTab & TAT-QA & FeTaQA & AIT-QA & TabMCQ & Avg. $\uparrow$ \\
    \midrule
    Upper bound (unpruned) & 2.59 &  & 88.9 & 33.1 & 45.1 & 46.8 & 33.0 & 43.6 & 61.3 & 50.3 \small \color{ForestGreen}{(100\%)} \\ 
    \midrule
    \rowcolor{lightblue}
    Ours (both $\mathcal{L}_p$, $\mathcal{L}_r$) & 1.78&\checkmark & 88.5 & 32.1 & 40.9 & 44.8 & 32.9 & 41.9 & 57.8 & 48.4 \small \color{ForestGreen}{(96.2\%)}\\
    \ \ - without $\mathcal{L}_p$ & 1.78&  \checkmark & 87.9 & 33.0 & 40.6 & 42.9 & 32.7 & 40.1 & 54.2 & 47.3 \small \color{ForestGreen}{(94.0\%)}\\
    \ \ \ \ \ - without $\mathcal{L}_r$ & 1.78& \checkmark & 76.9 & 25.0 & 22.2 & 31.0 & 25.5 & 24.9 & 55.1 & 37.2 \small \color{ForestGreen}{(74.0\%)} \\
    \midrule
    \end{tabular}
    \end{adjustbox}
    \caption{\textbf{Ablation on token focusing loss.}
    $\checkmark$ indicates pruning applied for inference. 
    Percentages are relative to the unpruned baseline (first row).
    }
    \label{tab:contrastive_token_sup_ablation_full_tab}
\end{table*}
\begin{table*}[t!]
    \centering
    \begin{adjustbox}{width=\textwidth}
    \begin{tabular}{c|c|ccccccc|cc}
    \toprule
      Training Loss & Token set for inference & TABMWP & WTQ & HiTab & TAT-QA & FeTaQA & AIT-QA & TabMCQ & Avg. & $\Delta$ \\
    \midrule
    \multirow{2}{*}{$\mathcal{L}_r$} & $\mathbf{V}_r$  & 87.9 & 33.0 & 40.6 & 42.9 & 32.7 & 40.1 & 54.2 & 47.3 & \multirow{2}{*}{33.8} \\
     & $\mathbf{V}_p$  & 16.9 & 8.8 & 2.5 & 0.5 & 15.4 & 0.2 & 50.4 & 13.5 & \\
    \midrule
    \multirow{2}{*}{$\mathcal{L}_r, \mathcal{L}_p$} & $\mathbf{V}_r$  & 88.5 & 32.1 & 40.9 & 44.8 & 32.9 & 41.9 & 57.8 & 48.4 & \multirow{2}{*}{48.4} \\
     & $\mathbf{V}_p$  & 0.0 & 0.0 & 0.0 & 0.0 & 0.0 & 0.0 & 0.0 & 0.0 &  \\
    \bottomrule
    \end{tabular}
    \end{adjustbox}
    \caption{\textbf{Performance comparison using $\mathbf{V}_r$ and $\mathbf{V}_p$ for inference.}
    $\Delta$ denotes the difference in average performance between the retained set $\mathbf{V}_r$ and pruned set $\mathbf{V}_p$ for inference.
    }
    \label{tab:margin_by_pruning_loss_full_tab}
\end{table*}
\begin{table*}[t!]
    \centering
    \begin{adjustbox}{width=\linewidth}
    \begin{tabular}{c|cc|ccccccc|c}
    \toprule
      Pruning rate $p$ & TFLOPs $\downarrow$& Memory$\downarrow$ & TABMWP & WTQ & HiTab & TAT-QA & FeTaQA & AIT-QA & TabMCQ & Avg.$\uparrow$ \\
    \midrule
    0.0 & 2.59 \small \color{ForestGreen}{(100\%)}& 18.7G \small \color{ForestGreen}{(100\%)} & 88.9 & 33.1 & 45.1 & 46.8 & 33.0 & 43.6 & 61.3 & 50.3 \small \color{ForestGreen}{(100\%)} \\ 
    \midrule
    0.1 & 2.31 \small\color{ForestGreen}{(89\%)}& 15.3G \small \color{ForestGreen}{(82\%)} & 88.4 & 32.4 & 42.8 & 45.1 & 32.9 & 46.4 & 57.9 & 49.4 \small \color{ForestGreen}{(98\%)} \\  
    0.2 & 2.03 \small\color{ForestGreen}{(78\%)}& 12.4G \small \color{ForestGreen}{(66\%)} & 88.6 & 32.3 & 42.1 & 45.1 & 32.8 & 42.2 & 56.5 & 48.5 \small \color{ForestGreen}{(96\%)} \\ 
    \rowcolor{lightblue}
    0.3 & 1.78 \small\color{ForestGreen}{(68\%)}& 11.2G \small \color{ForestGreen}{(60\%)} & 88.5 & 32.1 & 40.9 & 44.8 & 32.9 & 41.9 & 57.8 & 48.4 \small \color{ForestGreen}{(96\%)} \\ 
    0.4 & 1.52 \small\color{ForestGreen}{(59\%)}& 11.1G \small \color{ForestGreen}{(59\%)} & 87.6 & 31.1 & 38.2 & 42.5 & 32.6 & 37.4 & 57.2 & 46.7 \small \color{ForestGreen}{(93\%)} \\ 
    0.5 & 1.28 \small\color{ForestGreen}{(49\%)}& 11.0G \small \color{ForestGreen}{(59\%)} & 86.0 & 28.8 & 31.8 & 39.4 & 31.6 & 36.8 & 53.9 & 44.0 \small \color{ForestGreen}{(88\%)} \\ 
    \bottomrule 
    \end{tabular}
    \end{adjustbox}
    \caption{\textbf{Analysis on pruning rate $p$.}
    Percentages are relative to the unpruned baseline (first row).
    }
    \label{tab:pruning_rate_analysis_full}
\end{table*}
\begin{table*}[t!]
    \centering
    \begin{adjustbox}{width=\textwidth}
    \begin{tabular}{l|ccccccc|c}
    \toprule
    Suppression Loss $\mathcal{L}_p$ & TABMWP & WTQ & HiTab & TAT-QA & FeTaQA & AIT-QA & TabMCQ &Avg.$\uparrow$ \\
    \midrule
    None & 87.9 & 33.0 & 40.6 & 42.9 & 32.7 & 40.1 & 54.2 & 47.3 \\ 
    Negative CE & 40.7 & 25.3 & 28.7 & 34.7 & 5.6 & 21.5 & 34.0 & 27.2 \\
    \rowcolor{lightblue}
    Uniform KL (Ours) & 88.5 & 32.1 & 40.9 & 44.8 & 32.9 & 41.9 & 57.8  & 48.4 \\ 
    \bottomrule
    \end{tabular}
    \end{adjustbox}
    \caption{\textbf{Ablation on loss choice for suppression.}
    }
    \label{tab:suppresion_loss_choice_tab_full}
\end{table*}
\subsection{Analysis on Suppression Loss Choice}
\label{app_sec:suppression_loss_analysis}
In Tab.~\ref{tab:suppresion_loss_choice_tab_full}, we compare different formulations of the suppression loss $\mathcal{L}_p$.
\q{Negative CE} loss, minimizing the probability of the correct answer using pruned tokens $\mathbf{V}_p$, performs worse than omitting the loss.
This is likely because deliberately answering incorrectly still requires understanding the correct answer, leading to unintended information retention.
In contrast, enforcing uniform predictions more effectively prevents $\mathbf{V}_p$ from retaining meaningful information.

\subsection{Latency Analysis}
\label{app_sec:latency_analysis}
In Tab.~\ref{tab:latency_comparison}, the number of tokens generated per second (\ie throughput) and the average accuracy of each model are reported.
TabFlash (1B) attains the highest throughput while maintaining superior accuracy, and TabFlash (3B) achieves the best accuracy with competitive throughput.
These results demonstrate the efficiency of TabFlash together with its strong performance.
All metrics are computed on a single NVIDIA RTX A6000 GPU, using a system equipped with two AMD EPYC 7763 64-Core CPUs.

\section{Detailed Experimental Settings}
In this section, we provide additional experimental details.
To support reproducibility, the core implementation of our TabFlash architecture is included in separate files.
The full codebase will be released upon acceptance.
\label{app_sec:detailed}
\subsection{Proprietary MLLMs}
\label{app_sec:proprietary_mllms}
For GPT-4V, we take results from Table-LLaVA~\cite{zheng2024multimodal}.
For GPT-4o, we experimented with GPT-4o-2024-05-13 version through the official API.
For Gemini, we experimented with gemini-2.5-pro through the official API.

\subsection{Detailed Configurations about Question Conditioning}
\label{app_sec:question_cond_configurations}
\begin{table*}[t!]
    \centering
    \begin{adjustbox}{width=0.8\textwidth}
    \begin{tabular}{l|cc|c}
    \toprule
    Injection layers & Interval& Layer indices & Avg. \\ 
    \midrule
    All (1-24)& 1 & [1, 2, ..., 24] & 43.4 \\ 
    Early (1-8) & 1& [1, 2, 3, 4, 5, 6, 7, 8] & 44.8 \\ 
    Mid (9-16) & 1& [9, 10, 11, 12, 13, 14, 15, 16] & 48.0 \\ 
    Late (17-24)& 1& [17, 18, 19, 20, 21, 22, 23, 24] & 48.8 \\ 
    \midrule
    Sparse & 2& [2, 4, 6, 8, 10, 12, 14, 16, 18, 20, 22, 24] & 48.0 \\ 
    Sparse & 3& [3, 6, 9, 12, 15, 18, 21, 24] & 48.6 \\ 
    Sparse & 4& [4, 8, 12, 16, 20, 24] & 49.0 \\ 
    \midrule
    \rowcolor{lightblue}
    Progressive (Ours)& Progressive & [11, 14, 17, 19, 21, 23]& 50.3\\
    \midrule
    \end{tabular}
    \end{adjustbox}
    \caption{\textbf{Details about question injection layers.}
    Layer indices of All (1-24) are omitted for conciseness.
    }
    \label{tab:supp_conditioning_details}
\end{table*}
In Tab.~\ref{tab:supp_conditioning_details}, we report the exact layer indices where question embeddings are injected.
Unlike previous methods that use a fixed conditioning interval, our approach progressively increases the frequency of injection as the layers deepen.
Specifically, conditioning is applied every four layers in the early stages, then every three and two layers, and finally to every layer toward the end.
This progressive strategy enables stable injection of question information into the ViT layers, leading to improved performance.

\section{More Visualizations of ViT $L_2$ Norms}
\begin{figure*}[t]
    \centering
    \includegraphics[width=0.8\linewidth]{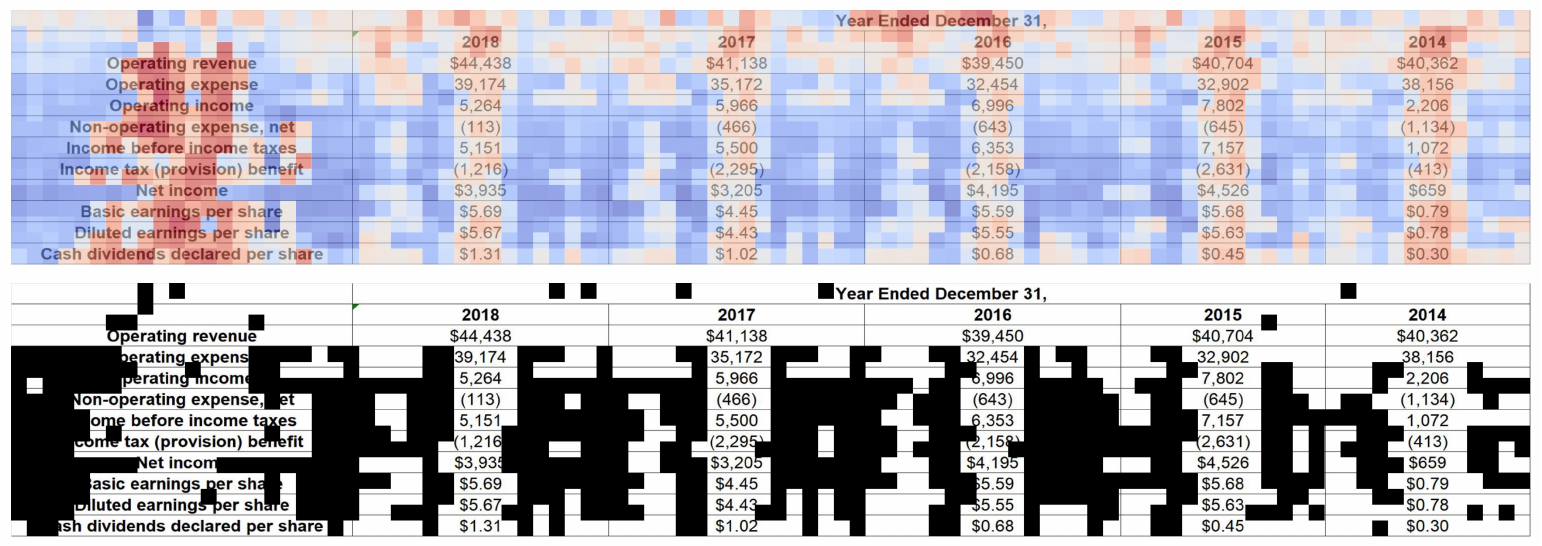}
    \includegraphics[width=0.5\linewidth]{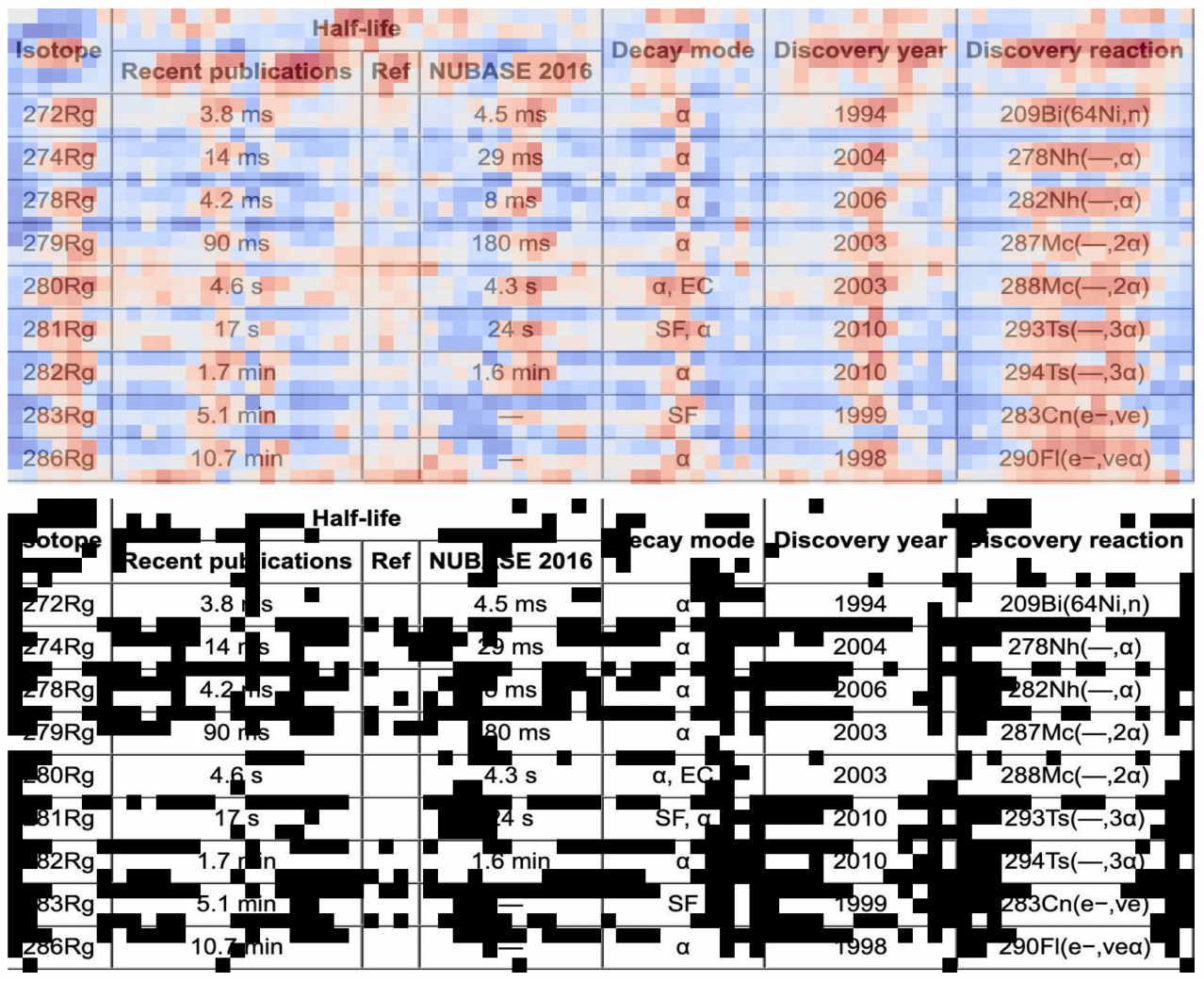}
    \includegraphics[width=0.5\linewidth]{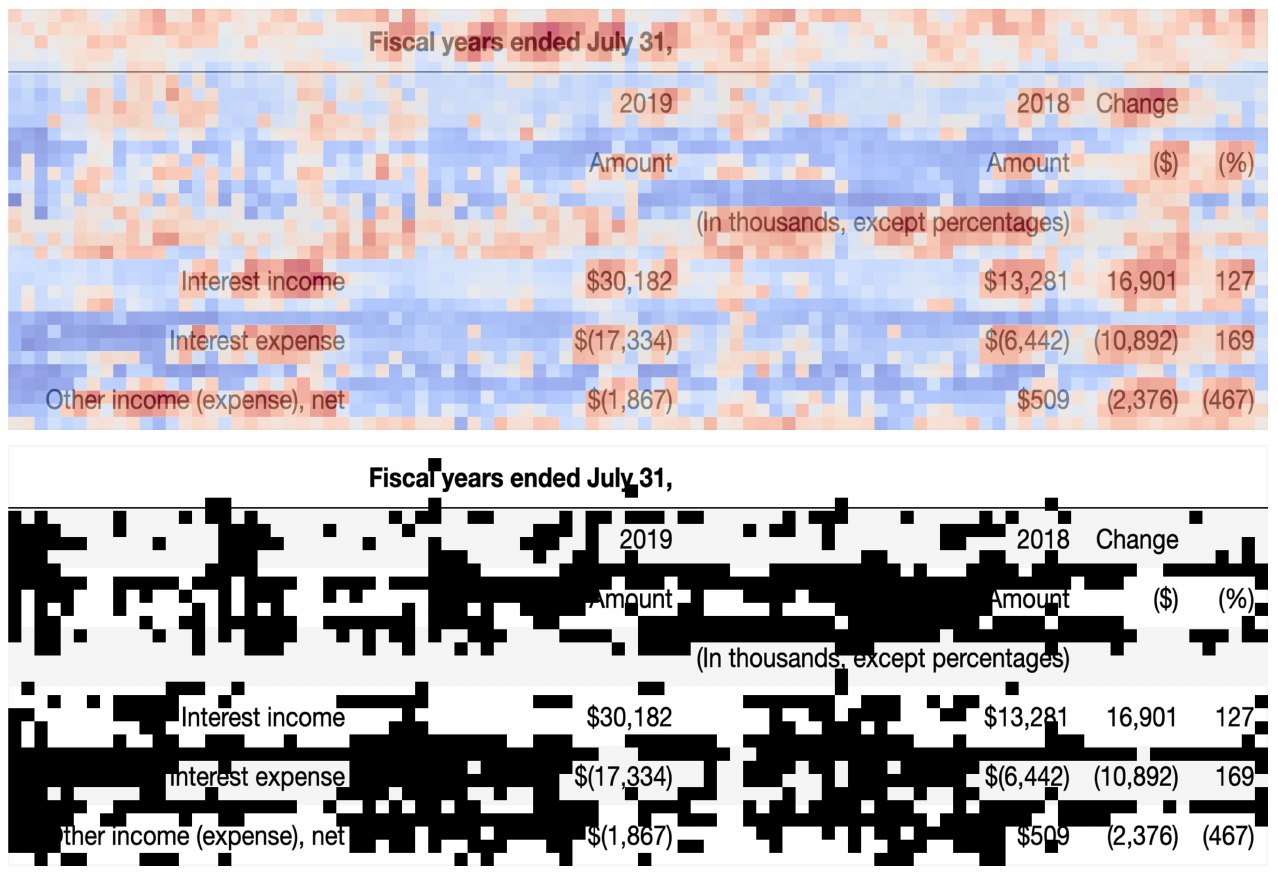}
    \caption{
    \textbf{Further visualization of $L_2$ norms of ViT output tokens (top) and norm-based pruning results (bottom).}
    \textcolor{red}{Red} and \textcolor{blue}{blue} color denotes \textcolor{red}{high} and \textcolor{blue}{low} $L_2$ norms, respectively.
    30\% of tokens with the lowest norms are pruned ($p=0.3)$.
    }
    \label{fig:supp_l2norm_qual}
\end{figure*}
\label{app_sec:l2norm}
In Fig.~\ref{fig:supp_l2norm_qual}, we provide additional visualizations of the $L_2$ norm of ViT output tokens and the results of norm-based pruning.
These visualizations confirm that the $L_2$ norm serves as an effective criterion for identifying background regions, allowing norm-based pruning to successfully remove redundant tokens and thereby enhance the efficiency of TabFlash.

\section{More Qualitative Results}
\begin{figure*}[t]
    \centering
    \includegraphics[width=\linewidth]{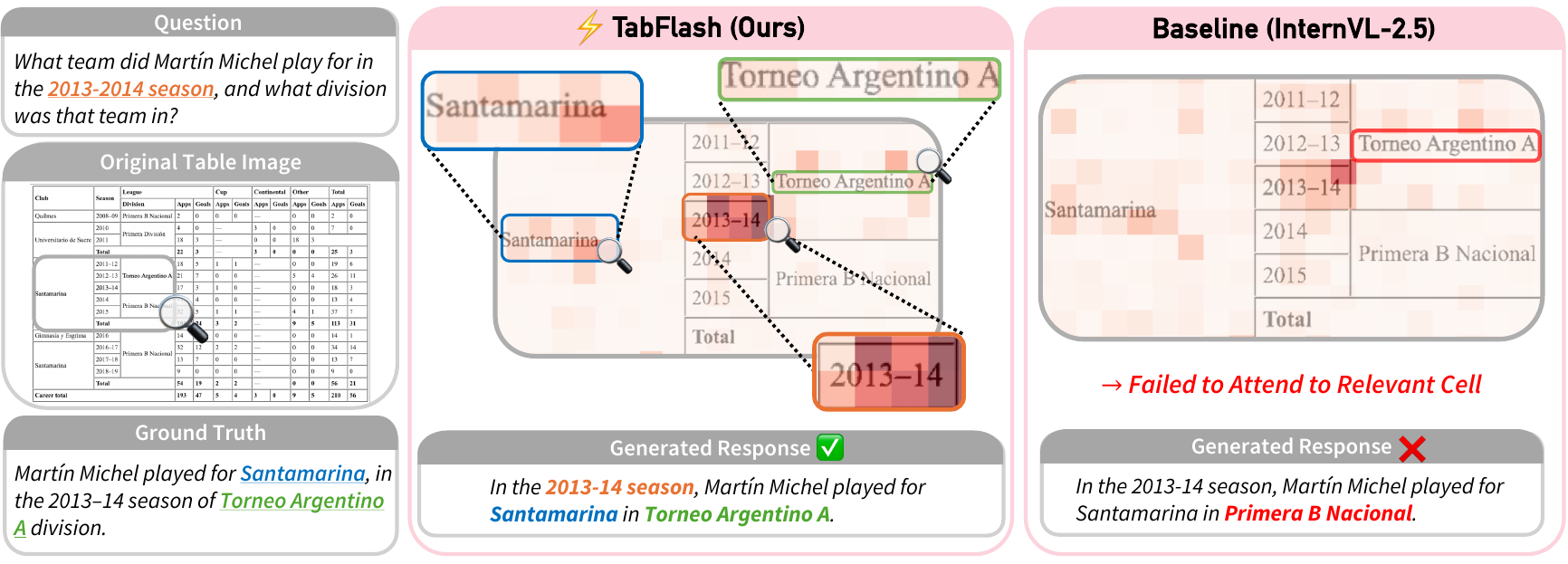}
    \includegraphics[width=\linewidth]{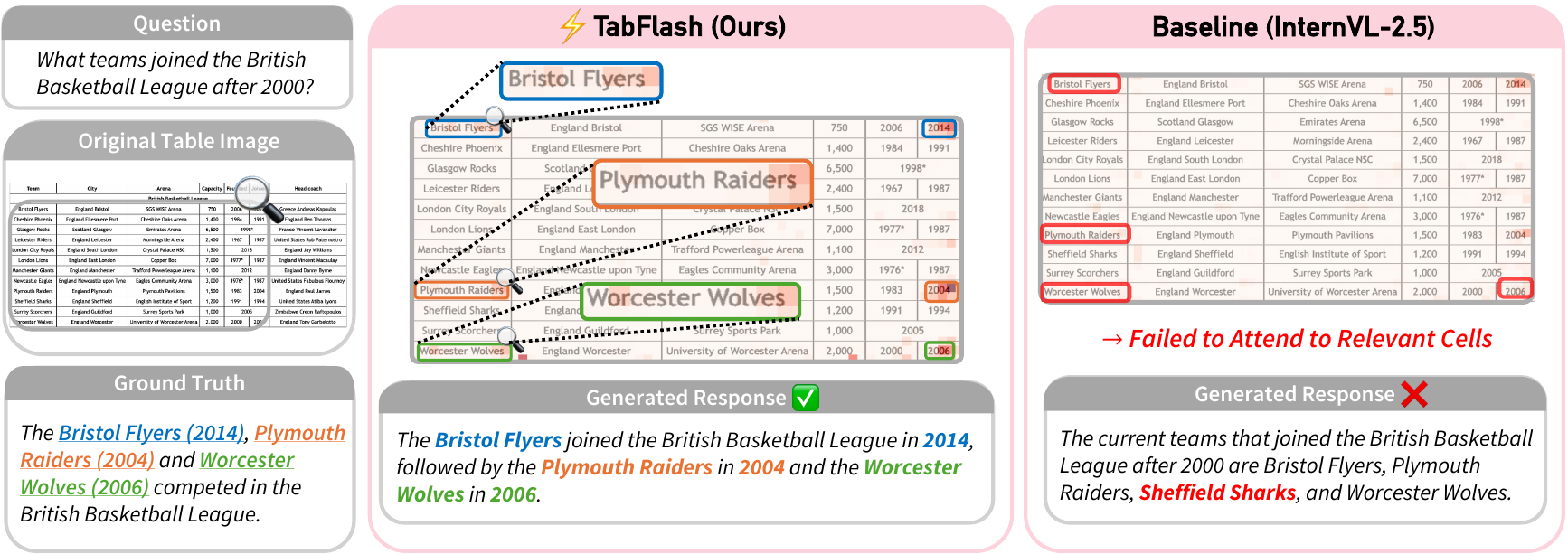}
    \includegraphics[width=\linewidth]{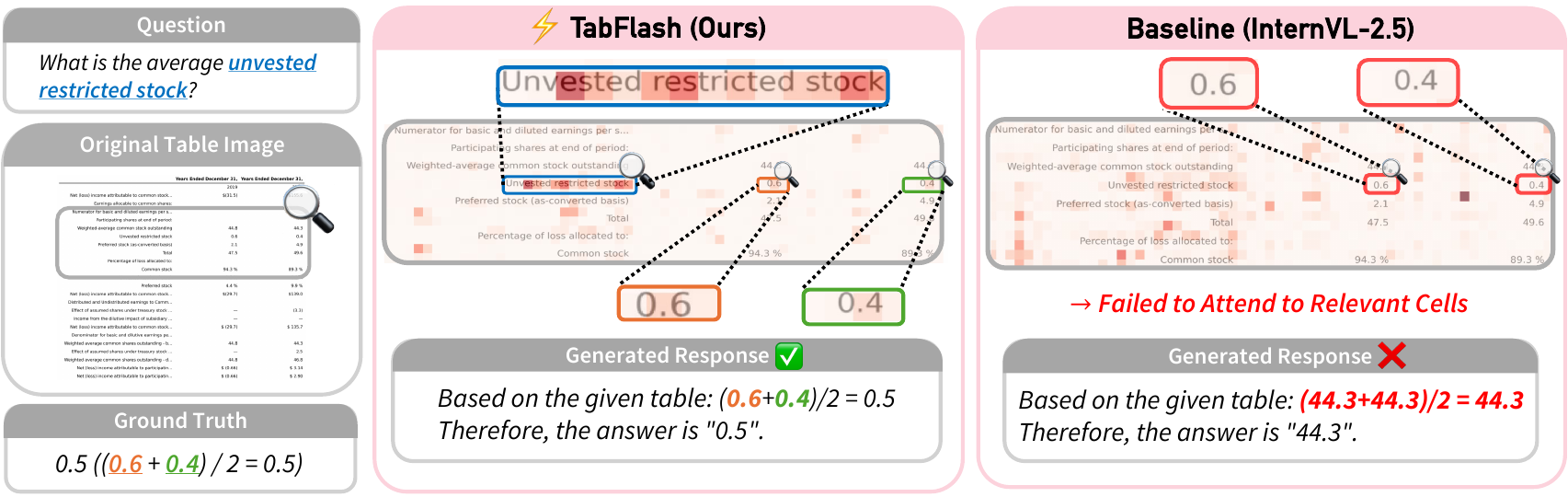}
    \caption{
    \textbf{Further qualitative examples.}
    Near-white regions indicate low attention, while stronger red colors represent higher attention scores.
    Best viewed when zoomed in.}
    \label{fig:supp_qual}
\end{figure*}
\label{app_sec:qual}
In Fig.~\ref{fig:supp_qual}, we present additional qualitative examples comparing attention scores on visual tokens and the generation results of TabFlash and InternVL-2.5, a baseline MLLM architecture.
The results demonstrate that TabFlash more effectively assigns higher attention to question-relevant regions, leading to correct responses.
In contrast, InternVL-2.5 fails to focus attention on relevant areas, instead spreading it broadly across the image, which often results in incorrect answers.

\end{document}